\documentclass[review]{elsarticle}
\graphicspath{ {./figures/} }
\usepackage{hyperref}
\usepackage{float}
\usepackage{verbatim} 
\usepackage{apalike}
\restylefloat{figure}
\restylefloat{table}
\usepackage{amsmath}

\usepackage{color}
\usepackage{adjustbox}
\usepackage[T1]{fontenc}
\usepackage[utf8]{inputenc}
\usepackage{graphicx, float}
\usepackage{enumitem}
\usepackage[table,xcdraw]{xcolor}
\usepackage{float}
\usepackage{tgadventor}
\usepackage[section]{placeins}

\journal{IASIM24 Workshop}

\biboptions{authoryear}

\begin{document}
\begin{frontmatter}

\title{Hyperspectral Dataset and Deep Learning methods for Waste from Electric and Electronic Equipment Identification (WEEE)}

\author[Tecnalia_address,ehu_address]{Artzai Picon\corref{cor1}}
\ead{artzai.picon@tecnalia.com}

\author[Tecnalia_address]{Pablo Galan}
\author[Tecnalia_address]{Arantza Bereciartua-Perez}
\author[Tecnalia_address]{Leire Benito-del-Valle}

\cortext[cor1]{Corresponding author.}
\address[Tecnalia_address]{TECNALIA, Basque Research and Technology Alliance (BRTA), Parque Tecnol\'{o}gico de Bizkaia, C/ Geldo. Edificio 700, E-48160 Derio - Bizkaia (Spain)}
\address[ehu_address]{University of the Basque Country, Plaza Torres Quevedo, 48013 Bilbao (Spain) }

\begin{abstract}

Hyperspectral imaging, a rapidly evolving field, has witnessed the ascendancy of deep learning techniques, supplanting classical feature extraction and classification methods in various applications. However, many researchers employ arbitrary architectures for hyperspectral image processing, often without rigorous analysis of the interplay between spectral and spatial information. This oversight neglects the implications of combining these two modalities on model performance.

In this paper, we evaluate the performance of diverse deep learning architectures for hyperspectral image segmentation. Our analysis disentangles the impact of different architectures, spanning various spectral and spatial granularities. Specifically, we investigate the effects of spectral resolution (capturing spectral information) and spatial texture (conveying spatial details) on segmentation outcomes. Additionally, we explore the transferability of knowledge from large pre-trained image foundation models, originally designed for RGB images, to the hyperspectral domain.

Results show that incorporating spatial information alongside spectral data leads to improved segmentation results, and that it is essential to further work on novel architectures comprising spectral and spatial information and on the adaption of RGB foundation models into the hyperspectral domain.

Furthermore, we contribute to the field by cleaning and publicly releasing the Tecnalia WEEE Hyperspectral dataset. This dataset contains different non-ferrous fractions of Waste Electrical and Electronic Equipment (WEEE), including Copper, Brass, Aluminum, Stainless Steel, and White Copper, spanning the range of 400 to 1000 nm.

We expect these conclusions can guide novel researchers in the field of hyperspectral imaging.

\end{abstract}

\begin{keyword}
Hyperspectral Imaging \sep WEEE \sep Recycling \sep Metal Scrap \sep Deep Learning
\end{keyword}

\end{frontmatter}


\section{Introduction}

In the constantly evolving field of hyperspectral imaging, deep learning techniques are replacing classical feature extraction and subsequent classification methods in some applications. However, many researchers use arbitrary architectures for hyperspectral image processing with little or no analysis of the influence of spectral and spatial information and its implications without considering the effect of their combination on model performance (\cite{picon2009fuzzy}). 

While foundational models have proven to be very effective in the RGB imaging domain, their application in the hyperspectral domain remains limited. Hyperspectral images, which capture a broad spectrum of light, provide information beyond what can be seen in RGB images. However, the unique characteristics of hyperspectral data, such as high dimensionality and spectral variability, pose new challenges that foundational models, trained primarily on RGB images, may not be equipped to handle.

In this paper we analyze the performance of different deep learning architectures for hyperspectral image segmentation. On the one hand, we analyze the effect of different architectures covering different spectral and spatial granularities to disentangle the effect of spectral information (spectral level resolution) and also spatial information containing image texture information. Additionally, we also measure the effect of fitting large pre-trained image foundation models (\cite{oquab2023dinov2}) on RGB images to hyperspectral images to see whether the knowledge extracted from RGB images can be efficiently transferred into the hyperspectral domain.

The objective of the paper is to show the current limitations of integrating common RGB architectures for hyperspectral imaging and to define the next steps in the application of deep learning technologies for hyperspectral data. 

In addition, we also clean and make available to the public the Tecnalia WEEE Hyperspectral dataset. This dataset contains different non-ferrous fractions of Waste Electrical and Electronic Equipment (WEEE) from Copper, Brass, Aluminum, Stainless Steel and White Copper in the range of 400 to 1000 nm.

\section{Dataset description}

Tecnalia Hyperspectral Dataset (\cite{picon2010automation}) contains different non-ferreous fractions of Waste from Electric and Electronic Equipment (WEEE) of Copper, Brass, Aluminum, Stainless Steel and White Copper. Images were captured by a hyperspectral Specim PHF Fast10 camera that is able to capture wavelengths in the range of 400 to 1000 nm with a spectral resolution of less than 1 nm. The PHF Fast10 camera is equipped with a CMOS sensor (1024 × 1024 resolution), a Camera Link interface and a special Fore objective OL10. The provided dataset contains 76 uniformly distributed wave-lengths in the spectral range [415.05 nm, 1008.10 nm]. Illumination setup, as described in \cite{picon2012real}, was specifically designed to reduce the specular reflections generated by the surface of the non-ferrous materials and to provide a homogeneous and even illumination that covers the wavelengths sensitive to the hyperspectral camera. The illumination system consists of a parabolic surface that uniformly distributes the light generated by 9 halogens and 18 white LEDs covering the spectral range between 400 to 1000 nm.

Acquired images were calibrated by using a white and dark Spectralon (\cite{bruegge1993use}) reference patterns to calculate the reflectance according to (\ref{eq:spectral_reflectance}), where $L(\lambda)$ is the observed intensity for each pixel, and $R_{BS}$ and $R_{BW}$ are the measured reflectances for the black and white Spectralon patterns, respectively.   

\begin{equation}
\label{eq:spectral_reflectance}
{Reflectance(\lambda) = \frac{L(\lambda) - R_{BS}(\lambda)}{R_{WS}(\lambda) - R_{BS}(\lambda)} } 
\end{equation}

\begin{figure*}[h]
    \centering
    \includegraphics[width=8.5cm]{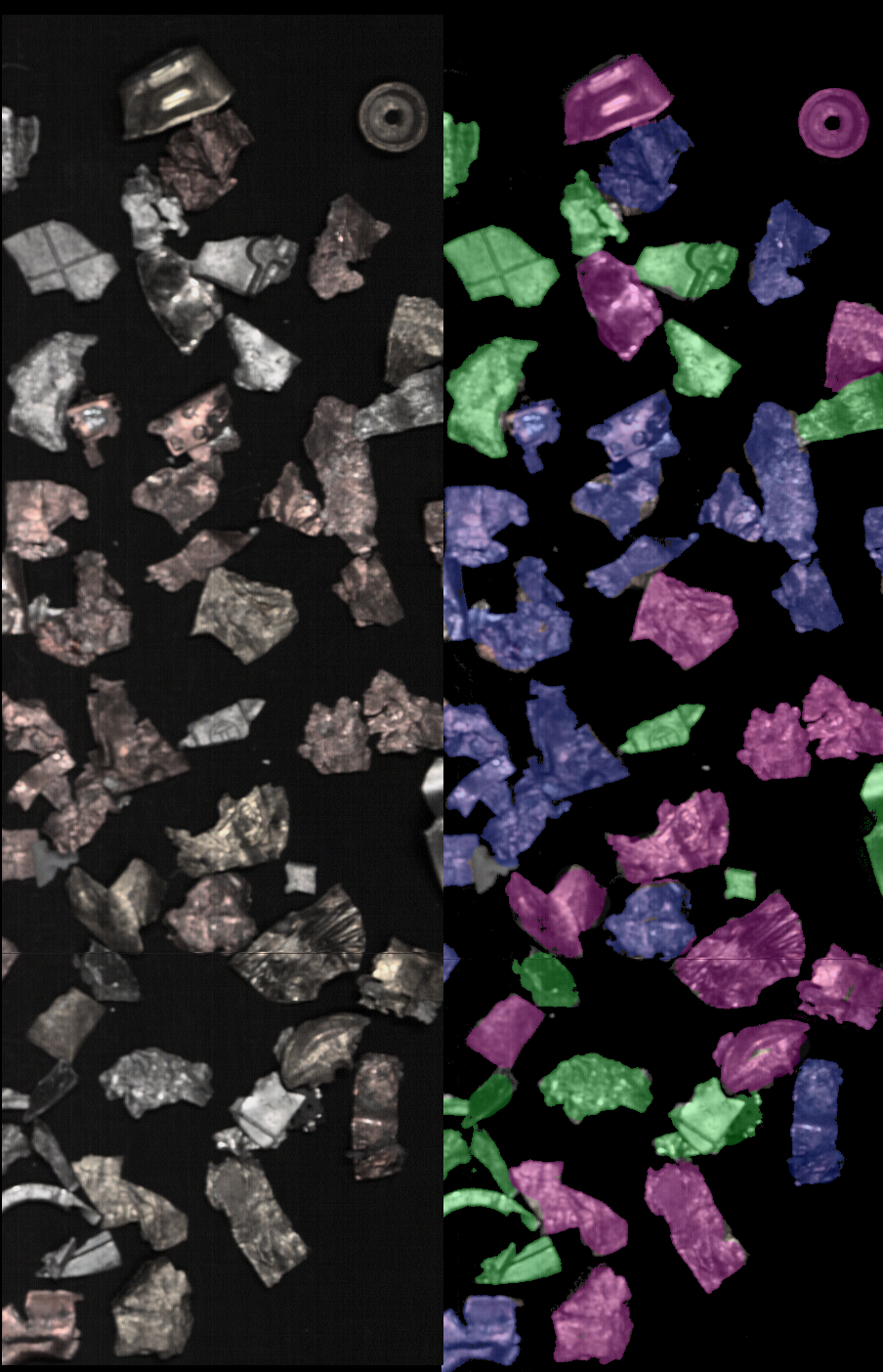}
    \caption{RGB and Mask representation of one dataset image, containing mixed fractions of copper, aluminum and brass.}
    \label{fig:dataset_example}
\end{figure*}

An example of an image from the dataset is provided in figure \ref{fig:dataset_example}. Three different materials can be appreciated, copper, brass and aluminum. The dataset consists of 13 independent images containing Copper, Brass, Aluminum, Stainless Steel and White Copper non-ferrous fractions of WEEE waste.

\section{Hyperspectral Architectures}
\label{sec:architectures}

\subsection{Basic architectures}
\begin{figure*}[h]
    \centering
    \includegraphics[width=12cm]{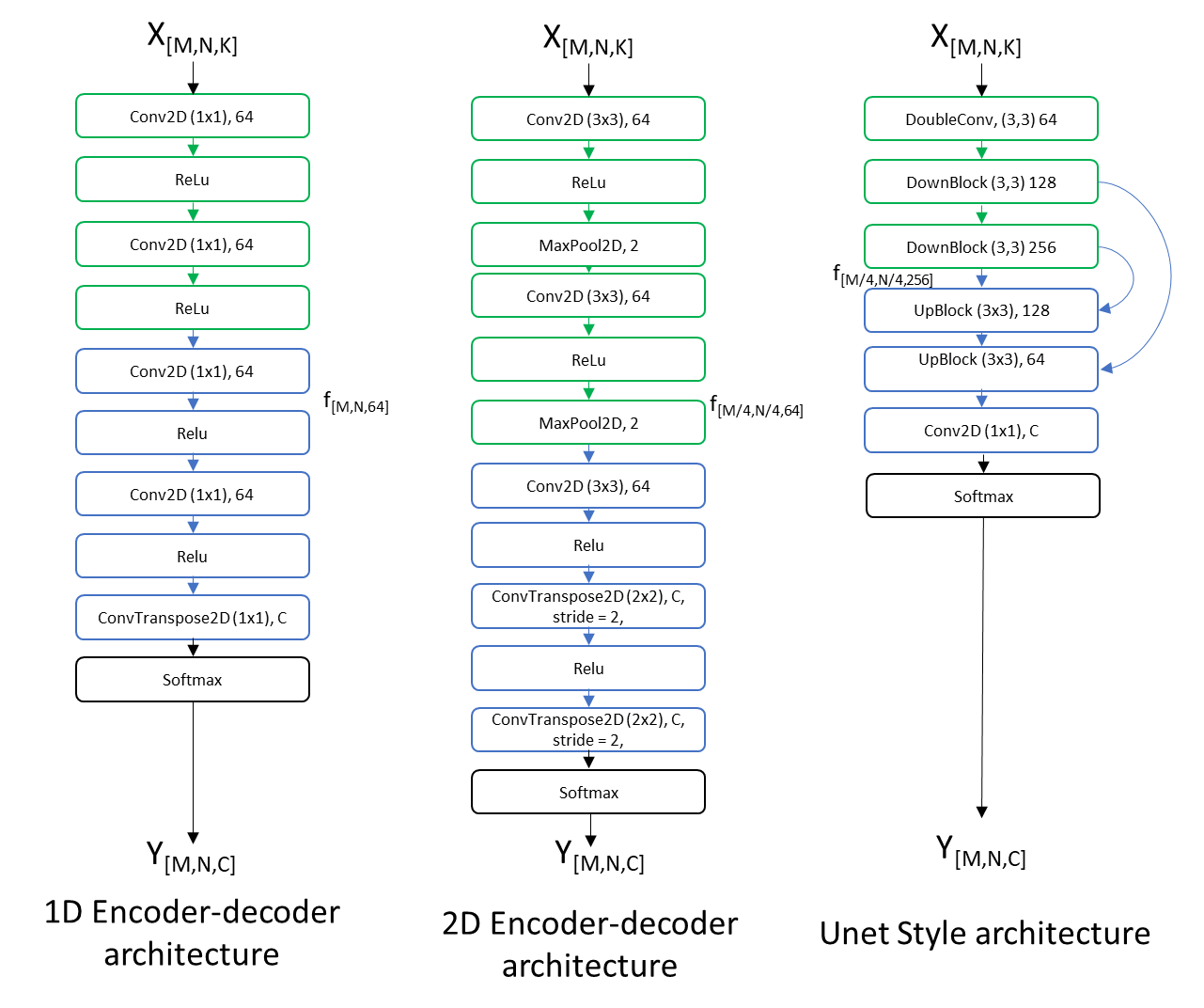}
    \caption{left) 1D encoder-decoder architecture, middle) 2D encoder-decoder architecture, right) U-Net Style architecture}
    \label{fig:architectures}
\end{figure*}

In this section, we describe the three basic architectures we will employ to analyse the effect of the spectral (wavelength) and the spatial (2D textures and patterns) on the WEEE recycling segmentation task. All architectures are depicted in figure \ref{fig:architectures}. These architectures have been trained from scratch, with no initial pre-trained weights.

\subsubsection{1D Encoder-Decoder architecture}
This architecture is composed of a set of convolutional filters of size (1x1) that are calculated along all the spectral axes (\cite{kuester20211d}). This means that the spatial information is not used and only information from each pixel is used for the segmentation task.

\subsubsection{2D Encoder-Decoder architecture}

This architecture is inspired by the Fully Convolutional Segmentation Neural Network (FCN) (\cite{long2015fully}). In this approach a VGG style (\cite{simonyan2014very}) encoder is used, composed by a set of convolutional filters to encode higher level features followed by max-pooling operations to reduce the spatial information. After these encoding operations, a decoding operation composed of a set of convolutional filters and transposed convolutional filters is used to recover the spatial information. With this approach, we encode both, the spectral information and the spatial information in a sole network.

\subsubsection{U-Net Style architecture}

In this approach, we extend the 2D encoder-decoder architecture by incorporating skip connections between the encoder and the decoder, which yields a more precise segmentation and, in addition, helps in reducing the number of required training images (\cite{ronneberger2015u}).

\subsection{Foundational Model Adapted architectures}
\label{ssec:foudnation_archi}

In order to evaluate whether the information captured from self-supervised large visual models over RGB images can be transferred into the hyperspectral domain, we will also evaluate the use of these models on the hyperspectral domain. To this end, we have taken a DINOv2 RGB architecture (\cite{oquab2023dinov2}) and extended it into the hyperspectral domain by making a slight change to the existing architecture. We simply duplicate the RGB weights from the pretrained patch embed layer, one channel at a time, until we obtain the same number of hyperspectral channels as the dataset. This layer is responsible for preprocessing the images by dividing them into fixed-size patches, which are then linearly embedded. As a result, the DINOv2 will require either 3, 7, or 76 input channels, depending on the experiment.  Besides this, we have concatenated an FCN-based decoder (\cite{long2015fully}) or a more modern Segformer decoder (\cite{xie2021segformer}), that presents more similarities with the U-Net decoder.

\section{Results}
\label{sec:results}

All the proposed architectures were trained on the Tecnalia Computing Platform, which is composed of several nodes with H100 NVIDIA graphic cards. Experiments were trained for 2000 epochs. 
In the case of the foundational based model, fine-tuning was performed for the first 1000 epochs and, for the other 1000 epochs, all weights were unfrozen. 
AdamW optimizer was selected and a learning rate of$10^-5$ was used. In order to measure the effect of the spectral bands, each architecture was tested against three spectral band configurations: 1) with all 76 bands, 2) selecting uniformly distributed bands, and 3) using the three bands that correspond to the Red, Green, and Blue color frequencies.

\subsection{Base models}

\begin{figure*}[h]
    \centering
    \includegraphics[width=12.5cm]{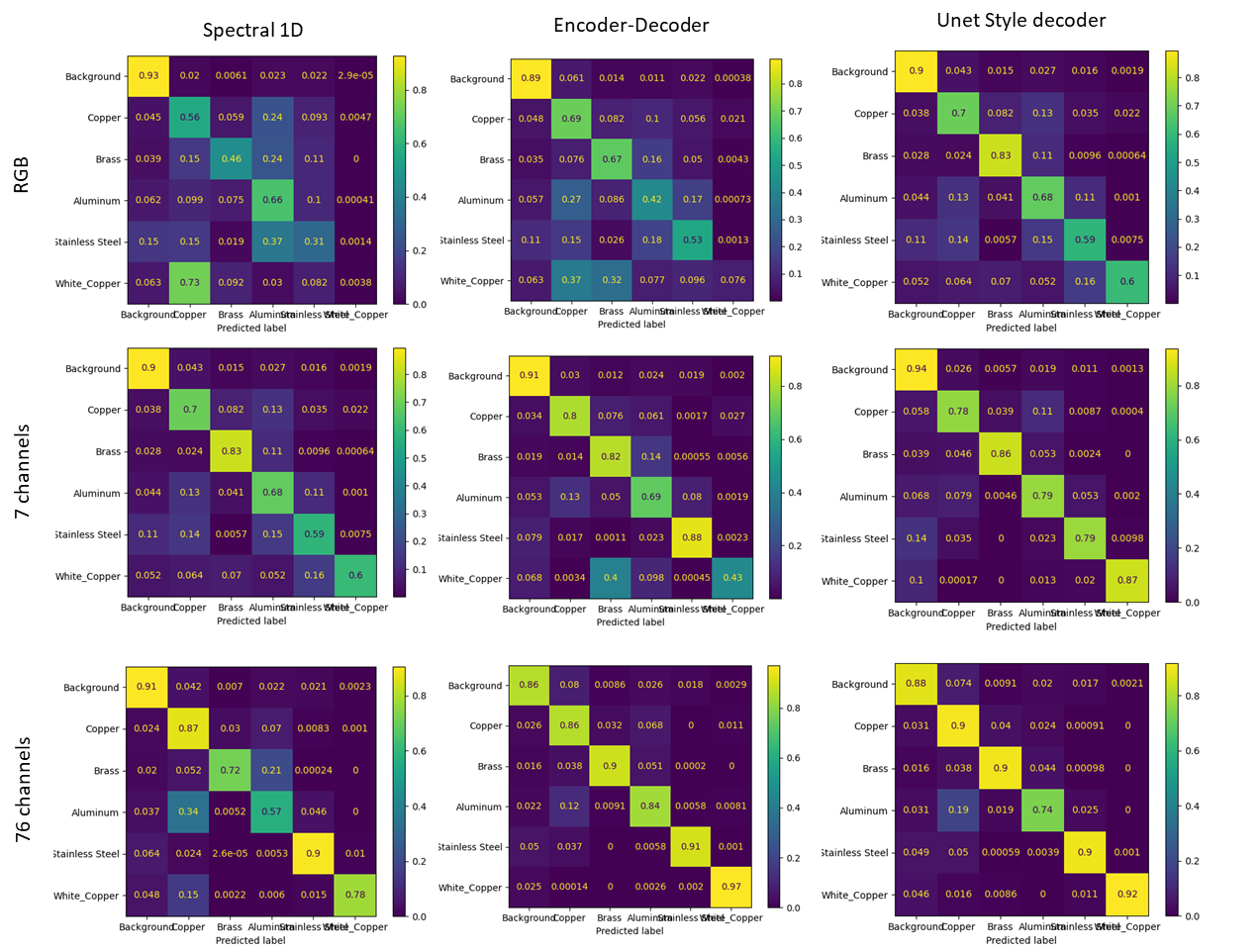}
    \caption{Confusion matrices over test set for the different experiments}
    \label{fig:cm}
\end{figure*}

\begin{figure*}[ht]
    \centering
    \includegraphics[width=12.5cm]{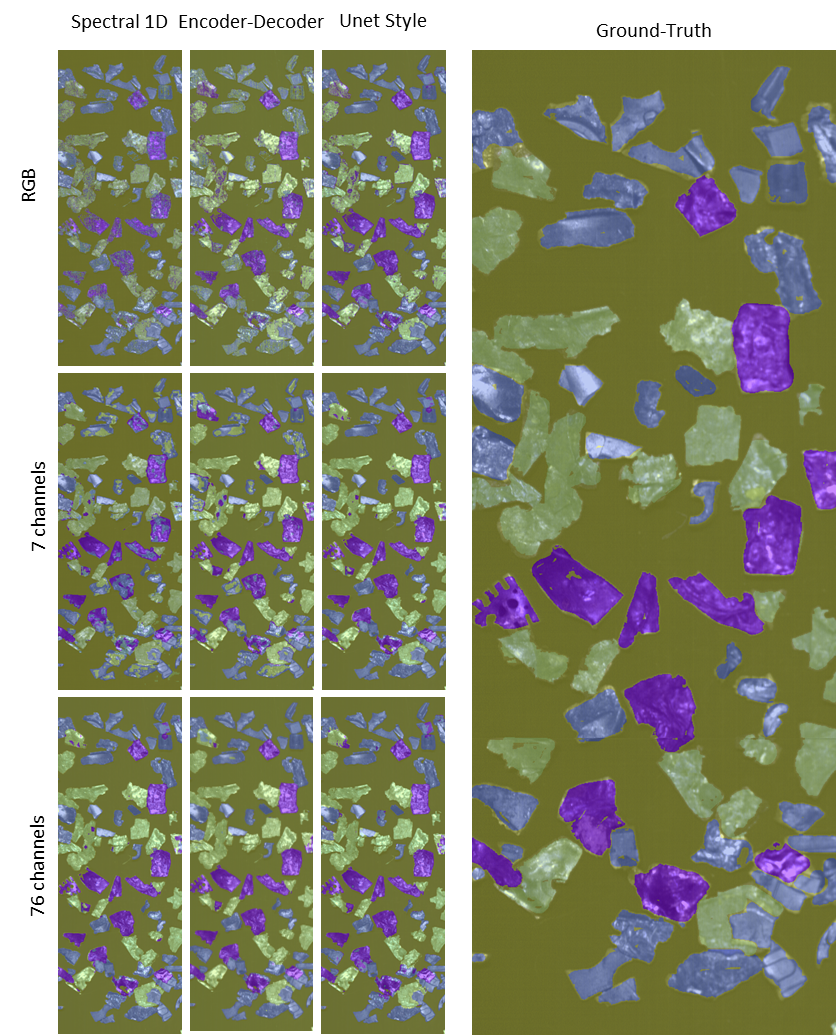}
    \caption{Predicted results over one test image for the different experiments and corresponding ground-truth}
    \label{fig:predictions}
\end{figure*}

Detailed results for the base architectures are depicted in table \ref{tab:results} whereas the summary of the results showing the mIoU of the models is depicted in table  \ref{tab:results_summary}. As it can be appreciated, the increase of both spatial information (2D decoders) and spectral information (more spectral bands) leads to a better performance of the model. This can also be appreciated in the confusion matrices over the test set (figure \ref{fig:cm}) and the example predictions (figure \ref{fig:predictions}).

\begin{table}[h]
  \centering
    \begin{tabular}{llll}
\hline
 & Only Spectral 1D & Encoder-Decoder & U-Net  \\
\hline
RGB bands & 0.36 & 0.49 & 0.64 \\
7 bands & 0.38 & 0.59 & \textbf{0.74} \\
76 bands & 0.55 & 0.72 & 0.73 \\
\hline
\end{tabular}
    \caption{mIoU metrics for the performed experiments.}
    \label{tab:results_summary}
\end{table}

If we analyse the spectral models that only use spectral information and no spatial one (experiments a1, a2, and a3), it can be seen that the performance improves when adding more spectral bands, going from a mIoU of 0.36 when using only RGB to 0.55 when using all of them. 

If we analyse the decoder type, it can be appreciated that adding architectures that can integrate spatial information together with the spectral one increases performance regardless the number of bands used in the model. When using RGB (experiments a1,b1,c1) we increase the mIoU from 0.36 when using only spectral information to 0.64 when using a U-Net style decoder. This phenomena is also enhanced when using 7 equally distributed bands (experiments a2,b2,c2). In this case, the mIoU is increased from 0.38, when using only pixel information to 0.74 when combining the U-Net style decoder. When using all bands (experiments a3,b3,c3), the addition of spatial decoders also contributes to better performance (mIoU from 0.55 to 0.73).

It is noticeable that the use of all bands together with the integration of spatial information does not lead to better results. This might be caused by the use of 2D convolutional filters that generate kernel matrices of size [Mf,Nf,K], where Mf and Nf are the filter spatial size and K are the numbers of bands of the model. This means that the spectral-wise filter generated by the network does not incorporate spectral neighborhood nor spectral-wise maxpooling operations to deal with spectral relationships being more prone to overfitting. It is necessary then, to develop channel-wise convolutions and architectures that can cope with the spectral filtering and to integrate them into hyperspectral deep learning architectures.
 
\begin{table}[h]
  \centering
  \resizebox{\textwidth}{!}{%
    \begin{tabular}{lllllllll}
\hline
Experiment & Spectral Bands & Backbone & Decoder & Precision & Recall & F1-Score & Accuracy & mIoU \\
\hline
a1 & 3 & VGG-style & spectral only & 0.49 & 0.49 & 0.48 & 0.92 & 0.36 \\
a2 & 7 & VGG-style & spectral only & 0.55 & 0.55 & 0.51 & 0.91 & 0.38 \\
a3 & 76 & VGG-style & spectral only & 0.69 & 0.71 & 0.69 & 0.93 & 0.55 \\
b1 & 3 & VGG-style & Encoder-decoder & 0.68 & 0.67 & 0.65 & 0.92 & 0.49 \\
b2 & 7 & VGG-style & Encoder-decoder & 0.72 & 0.75 & 0.73 & 0.95 & 0.59 \\
b3 & 76 & VGG-style & Encoder-decoder & \textbf{0.84} & 0.84 & \textbf{0.84} & \textbf{0.96} & 0.72 \\
c1 & 3 & VGG-style & U-Net & 0.78 & 0.79 & 0.77 & 0.94 & 0.64 \\
c2 & 7 & VGG-style & U-Net & 0.81 & \textbf{0.89 }& \textbf{0.84} & 0.95 & \textbf{0.74} \\
c3 & 76 & VGG-style & U-Net & 0.83 & 0.87 & 0.84 & 0.95 & 0.73 \\
\hline
\end{tabular}
  }
    \caption{Metrics for the performed experiments.}
    \label{tab:results}
\end{table}

\subsection{Adapted foundational models}

Additionally, we have analysed the adaptation of foundational models described in section \ref{ssec:foudnation_archi} with the two proposed encoders for the different numbers of bands. Results are depicted in table \ref{tab:results_foundation}. We appreciate that our results show that there is no benefit for this dataset from the use of RGB-adapted foundational models. However, the analysis of this is beyond this work as specific fine-tuning techniques or strategies such as LoRa (\cite{hu2021lora}) should be used.

\begin{table}[h]
  \centering
  \resizebox{\textwidth}{!}{%
    \begin{tabular}{lllllllll}
\hline
Experiment & Spectral Bands & Backbone & Decoder & Precision & Recall & F1-Score & Accuracy & mIoU \\
\hline
d1 & 3 & dinov2-small & Encoder-decoder & 0.51 & 0.5 & 0.5 & 0.9 & 0.36 \\
d2 & 7 & dinov2-small & Encoder-decoder & 0.81 & 0.87 & 0.84 & 0.96 & 0.72 \\
d3 & 73 & dinov2-small & Encoder-decoder & 0.77 & 0.84 & 0.79 & 0.94 & 0.66 \\
e1 & 3 & dinov2-small & U-Net & 0.73 & 0.8 & 0.76 & 0.94 & 0.62 \\
e2 & 7 & dinov2-small & U-Net & 0.82 & 0.8 & 0.8 & 0.95 & 0.67 \\
e3 & 73 & dinov2-small & U-Net & 0.79 & 0.83 & 0.81 & 0.95 & 0.68 \\
\hline
\end{tabular}
  }
    \caption{Metrics for the performed experiments.}
    \label{tab:results_foundation}
\end{table}

\section{Conclusions}
\label{sec:conclussions}

In this paper, we have publicly released the Tecnalia Hyperspectral WEEE dataset and investigated the impact of various base deep learning architectures on hyperspectral image segmentation.

Our findings confirm that incorporating spatial information alongside spectral data leads to improved segmentation results. However, due to the encoding of spectral information through classic 2D convolutions covering the entire spectral range, efficient processing through novel architectures is essential for better integration of spectral and spatial features.

Furthermore, foundational models have become indispensable in today’s RGB computer vision applications. To advance hyperspectral imaging, it is crucial to explore novel approaches for integrating and adapting these foundational models specifically for hyperspectral data.

We expect that these conclusions will inspire hyperspectral imaging researchers to develop innovative architectures and fine-tuning techniques, ultimately enhancing hyperspectral image segmentation capabilities.

\section{Data Availability Statement}
\label{sec:data_availability}
The data analyzed in this paper has been published in the Zenodo repository: \href{https://zenodo.org/records/12565131}{Zenodo Tecnalia Hyperspectral WEEE dataset}. It is available for non-commercial purposes.

Code is uploaded to GitHub at: \href{https://github.com/samtzai/tecnalia_weee_hyperspectral_dataset}{GitHub Repository}.

\section*{Funding}
Some authors have received support by the Elkartek Programme, Basque Government (Spain) (SMART-EYE (KK-2023/00021).

This paper was created to support Deep Learning on Hyperspectral Imaging Workshop and course that took place in IASIM 2024, Bilbao. 

\section*{Author Contributions}

\textbf{AP}: conceptualization, investigation, software, formal analysis, methodology, and writing--original draft, review and editing. 
\textbf{PG}: conceptualization, formal analysis, investigation and methodology, review and editing. 
\textbf{AB-P}: conceptualization, dataset, investigation and methodology, review and editing. 
\textbf{L-BV}: conceptualization, formal analysis, investigation and methodology, review and editing. 




\section*{Declaration of competing interest}

Authors declare that the research was conducted in the absence of any commercial or financial relationships that could be construed as a potential conflict of interest.





\bibliography{template}

\end{document}